\documentclass[10pt, a4paper, logo]{googledeepmind}

\usepackage[T1]{fontenc}
\usepackage{times}
\usepackage{microtype}
\usepackage{textcomp}
\usepackage{xspace}

\usepackage{amsmath, amsfonts, amssymb}

\usepackage{amsmath,amsfonts,bm}
\usepackage{natbib}
\usepackage{fancyhdr}

\def\eqref#1{equation~\ref{#1}}

\def\1{\bm{1}}

\newcommand{\MI}{\mathrm{I}}

\def\va{{\bm{a}}}

\def\ve{{\bm{e}}}

\def\vh{{\bm{h}}}

\def\vx{{\bm{x}}}
\def\vy{{\bm{y}}}

\DeclareMathAlphabet{\mathsfit}{\encodingdefault}{\sfdefault}{m}{sl}
\SetMathAlphabet{\mathsfit}{bold}{\encodingdefault}{\sfdefault}{bx}{n}

\def\gA{{\mathcal{A}}}

\def\gG{{\mathcal{G}}}

\def\gL{{\mathcal{L}}}

\def\gP{{\mathcal{P}}}

\def\gX{{\mathcal{X}}}
\def\gY{{\mathcal{Y}}}

\def\bbE{{\mathbb{E}}}

\usepackage[table]{xcolor}
\usepackage{booktabs}
\usepackage{multirow}
\usepackage{array}
\usepackage{colortbl}
\usepackage{siunitx}

\usepackage{graphicx}
\usepackage{subcaption}
\usepackage[leftcaption]{sidecap}
\usepackage{wrapfig}
\usepackage{float}
\usepackage{stfloats}

\usepackage[ruled, linesnumbered, lined]{algorithm2e}
\usepackage{enumitem}

\usepackage[most]{tcolorbox}
\tcbuselibrary{breakable}
\usepackage[normalem]{ulem}
\usepackage{verbatim}

\usepackage{natbib}
\usepackage{url}
\usepackage{hyperref}

\usepackage{etoolbox}

\definecolor{pos}{HTML}{E6F4EA}     
\definecolor{neg}{HTML}{FEF4F4}     
\newcommand{\ours}{CLIPO\xspace}    
 
\renewcommand{\titlefont}{\color{black}\normalfont\bfseries\fontsize{19}{22}\selectfont}

\title{CLIPO: Contrastive Learning in Policy Optimization Generalizes RLVR}

\author[1,2]{Sijia Cui$^{*}$}
\author[1]{Pengyu Cheng\textsuperscript{\dag}}
\author[1]{Jiajun Song}
\author[1]{Yongbo Gai}
\author[1]{Guojun Zhang}
\author[1]{Zhechao Yu}
\author[1]{Jianhe Lin}
\author[1]{Xiaoxi Jiang}
\author[1]{Guanjun Jiang}
\affil[1]{Qwen Large Model Application Team, Alibaba}
\affil[2]{Institute of Automation, Chinese Academy of Sciences}

\begin{abstract}
Reinforcement Learning with Verifiable Rewards (RLVR) has significantly advanced the reasoning capacity of Large Language Models (LLMs).
However, RLVR solely relies on final answers as outcome rewards, neglecting the correctness of intermediate reasoning steps. Training on these process-wrong but outcome-correct rollouts can lead to hallucination and answer-copying, severely undermining the model's generalization and robustness. 
To address this, we incorporate a \textbf{C}ontrastive \textbf{L}earning mechanism \textbf{i}nto the \textbf{P}olicy \textbf{O}ptimization (CLIPO) to generalize the RLVR process.
By optimizing a contrastive loss over successful rollouts, CLIPO steers the LLM to capture the invariant structure shared across correct reasoning paths. This provides a more robust cross-trajectory regularization than the original single-path supervision in RLVR, effectively mitigating step-level reasoning inconsistencies and suppressing hallucinatory artifacts.
In experiments, CLIPO consistently improves multiple RLVR baselines across diverse reasoning benchmarks, demonstrating uniform improvements in generalization and robustness for policy optimization of LLMs. 
Our code and training recipes are available at \url{https://github.com/Qwen-Applications/CLIPO}.
\end{abstract}

\begin{document}
\maketitle


\begin{quote}
\textit{``Happy families are all alike; every unhappy family is unhappy in its own way.''}

\hfill--- Leo Tolstoy, \textit{Anna Karenina}
\end{quote}

\vspace{-8mm}
\section{Introduction}
\begin{wrapfigure}{r}{0.6\textwidth} 
    \vspace{-6.3mm}
    \centering
    \includegraphics[width=0.6\textwidth]{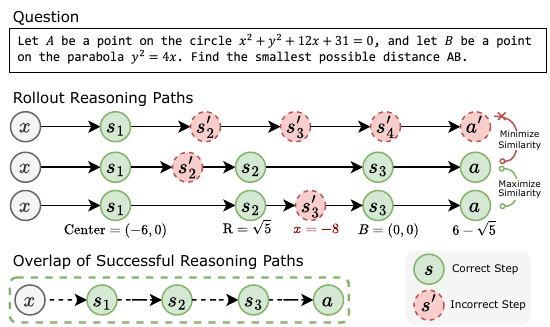}
    \caption{Intuition of \ours. Standard RLVR only relies on outcomes, neglecting the quality of intermediate reasoning steps. 
    \ours addresses this limitation by maximizing similarity between successful reasoning trajectories. 
    By aligning multiple positive rollouts, \ours identifies the invariant reasoning structure, \textit{i.e.},  the  “overlap” of successful paths, implicitly eliminating incorrect and hallucinative reasoning steps.
    }
    \label{fig:intro}
\end{wrapfigure}

Reinforcement learning with verifiable rewards (RLVR) has recently emerged as a mainstream paradigm for improving the reasoning capabilities of Large Language Models (LLMs)~\citep{jaech2024openai,guo2025deepseek,team2025kimi}.
Unlike Reinforcement Learning from Human Feedback (RLHF)~\citep{ouyang2022training,cheng2024adversarial,li2026dir}, which incurs substantial human annotation costs and suffers from inconsistent human judgment, RLVR leverages external environments (\textit{e.g.}, mathematical verifiers or code compilers) to provide objective, explicit, and consistent feedback. To train LLMs via RLVR, 
group-based policy optimization methods, such as Group Relative Policy Optimization (GRPO)~\citep{shao2024deepseekmath}, estimate relative advantages within a set of sampled responses and have demonstrated strong empirical performance across a wide range of reasoning tasks, including mathematics~\citep{yu2025dapo,zheng2025gspo,zhao2025gmpo,gao2025sapo}, coding~\citep{zhoubian2025rest,dai2026group}, deep search~\citep{jin2025searchr,lu2026ssp,wang2025vrag,sun2025zerosearch,zheng2025deepresearcher} and agents~\citep{chen2025reinforcement,gao2025beyond,dong2026agentic}.

However, RLVR faces ongoing doubts about its generalization capabilities~\citep{alam2025limits,chen2026exploration}, for it relies on outcome-based rewards without verifying intermediate reasoning steps, in which models are prone to overfitting by memorizing ground-truth answers~\citep{wu2025reasoning,ruan2025unveiling,yan2026spurious}. 
%
%
Such a coarse and binary rewarding scheme fails to distinguish between logically sound and spurious reasoning, as it collapses diverse reasoning paths sharing the same outcome into identical reward signals.
To address this weakness, a series of recent works have explored process reward models (PRMs) that provide finer-grained supervision by identifying errors during the reasoning process~\citep{lightman2023let,wang2024math,luo2024improve,zheng2025processbench,zhang-etal-2025-lessons}. The critical challenge for PRM methods lies in the substantial human annotation required to collect high-quality process reward data, making it costly and difficult to scale~\citep{zhang-etal-2025-lessons}.
Alternatively, some works leverage token-level entropy to provide fine-grained training guidance and enhance exploration~\citep{chen2025seed,zhang2025edge,tan2025gtpo,jiang2025rethinking,wang2025lambda,huang2026spotlight}. 
Nevertheless, such methods primarily reflect distributional uncertainty rather than semantic logical importance. Furthermore, their effectiveness is highly dependent on model capacity, often failing to provide reliable exploration signals as model scale varies~\citep{cui2025entropy,wang2025beyond}.



On the other hand, contrastive learning~\citep{chen2020simple,he2020momentum,radford2021learning} has been widely recognized as an unsupervised training scheme with strong generalization performance~\citep{saunshi2019theoretical,krishna2022rankgen,castricato2022robust}, in which models learn informative representations by maximizing similarity between positive pairs and minimizing it between negatives.
Since RLVR methods intrinsically utilize binary outcome rewards and target at enlarging the reward gap between positive (successful) and negative (failed) rollouts, an interesting question naturally arises: \textit{``Can contrastive learning enhance the generalization of RLVR?''} By designating successful and failed trajectories as positive and negative instances, contrastive learning can be seamlessly integrated into RLVR. 
Our insight, illustrated in Figure~\ref{fig:intro}, is that successful reasoning paths share a consistent underlying logic, whereas intermediate errors/hallucinations manifest as sporadic, uncorrelated noise.
By enforcing proximity among successful trajectories in the embedding space, contrastive learning acts as a denoising mechanism: it amplifies the invariant logical flow while suppressing non-systematic reasoning failures and hallucinations, ultimately leading to more robust generalization.
Based on this analysis, we propose {\textbf{C}ontrastive \textbf{L}earning \textbf{i}n \textbf{P}olicy \textbf{O}ptimization (\ours)}, a novel framework that integrates contrastive learning into group-based policy optimization. In \ours, a lightweight auxiliary head projects reasoning trajectories into an embedding space, where an InfoNCE objective~\citep{oord2018representation} is applied within each rollout group. This objective serves a dual purpose: it maximizes consistency among correct trajectories (positives) while enforcing a margin from erroneous ones (negatives). This alignment process compels the model to distill a shared logical essence across diverse successful rollouts, effectively disentangling it from faulty reasoning paths. The resulting contrastive loss is repurposed as a dense, auxiliary reward signal that complements the sparse, outcome-based feedback, providing a more informative gradient for policy optimization. 
%
We validate the effectiveness of \ours through two experimental tracks of increasing complexity.
In the first track, models are trained on the GSM8K dataset~\citep{cobbe2021training} and evaluated across 8 diverse benchmarks.
The second track involves training on MATH 7.5K~\citep{hendrycks2021measuring}, and evaluating on 6 challenging competition-level mathematics benchmarks.
Across both tracks, \ours consistently outperforms standard RLVR baselines.
Notably, our method shows substantial performance gains on perturbed and symbolic tasks, highlighting its robustness and generalization.

\vspace{-1mm}
\section{Preliminary}
\vspace{-1mm}
\subsection{Reinforcement Learning with Verifiable Rewards}
Unlike traditional reinforcement learning from human feedback (RLHF) relying on a learned reward model, RLVR~\citep{jaech2024openai,guo2025deepseek,team2025kimi} directly leverages deterministic environment feedback (e.g., unit tests for coding, or equivalence checkers for mathematics) to provide binary reward outcomes. 
Formally, given a prompt $\vx \in \mathcal{X}$, the policy model $\pi_\theta(\cdot|\vx)$ generates a response $\vy\in \gY$. The objective of RLVR is to maximize the expected verifiable reward: 
\begin{equation} \label{eq:rlvr_objective}
    \mathbb{E}_{\vx\sim \mathcal{X}, \vy \sim \pi_\theta(\cdot|\vx)} \big[ r(\vx, \vy) \big] - \beta \text{KL}[\pi_\theta \Vert \pi_{\text{ref}}],
\end{equation}
where $\pi_{\text{ref}}$ is a reference model and usually set as the initial checkpoint of $\pi_\theta$, $\text{KL}[\pi_\theta \Vert \pi_{\text{ref}}]$ is the KL divergence between the policy and reference model to maintain the training stability,  and $\beta>0$ is the coefficient for the KL penalty. Without loss of generality, we simplify the objective by omitting the KL penalty term in the following sections.
The input $\vx$ is associated with a ground-truth answer $\va^*$, and the verifiable reward $r(\vx,\vy)$ directly check if $\va^*$ and the predicted answer $\va = \gA(\vy)$ is equivalent: 
$r(\vx,\vy) = \mathbf{1}\{\gA(\vy) = \va^*\},$
where $\mathbf{1}\{\cdot\}$ is a binary indictor, and $\gA(\cdot)$ represents the predicted reasoning answer from the response $\vy$.

To optimize the above RLVR objective, policy gradient methods, such as Proximal Policy Optimization (PPO)~\citep{schulman2017proximal} and Group Relative Policy Optimization (GRPO)~\citep{shao2024deepseekmath}, dominate as the main solution. Generally, policy optimization methods estimate the gradient of \eqref{eq:rlvr_objective} with the REINFORCE estimator~\citep{williams1992simple}:
$\bbE_{\vx\sim \gX, \vy\sim\pi_\theta(\cdot|\vx)} \Big[ \sum_{t=1}^{|\vy|} \nabla_\theta \log(\pi_\theta(y^t| \vy^{<t}, \vx) \cdot  \hat{A}^t \Big],$
where $\vy^{<t} = (y_1,y_2,\dots,y_{t-1})$ is the length-$t$ prefix of the token sequence $\vy$, and $\hat{A}^t$ is estimated advantage of the $t$-th token $y^t$ based on the outcome reward $r(\vx,\vy)$. PPO deploys a critic model to predict each token's value and calculate the advantage $\hat{A}^t$ with Generalized Advantage Estimation (GAE)~\citep{schulman2015high}. GRPO
eliminates the need for the additional critic model by estimating the baseline within a group of sampled responses $\gG = \{\vy_1, \vy_2, \dots, \vy_G\} \sim \pi_\theta(\cdot|\vx)$ and their corresponding rewards $\{r_i = r(\vx,\vy_i)\}_{i=1}^G$. 
GRPO assumes all tokens within one rollout have the same advantage, which is computed by normalization over the rollout group:
$\hat{A}^t_i = \frac{r_i - \text{Mean}(r_1, r_2,\dots, r_G)}{\text{Std}(r_1, r_2, \dots, r_G)}$.
Therefore, the overall policy gradient for GRPO is:
\begin{equation}
 \bbE_{\vx\sim \gX}  \Big[  \frac{1}{G}\sum_{i=1}^G \Big[ \sum_{t=1}^{|\vy_i|} \nabla_\theta \log(\pi_\theta(y_i^t| \vy_i^{<t}, \vx) \cdot  \hat{A}_i^t \Big]\Big].
\end{equation}
%
%
To further improve GRPO's stability and scalability, Group Sequence Policy Optimization (GSPO)~\citep{zheng2025gspo} shifts optimization from token-level importance weighting to the sequence level, effectively mitigating variance in large-scale training. 
Meanwhile, Decoupled Clip and Dynamic sAmpling Policy Optimization (DAPO)~\citep{yu2025dapo} introduces asymmetric clipping bounds and dynamic sampling to maintain effective gradients in sparse reward environments.
Despite these algorithmic refinements, existing RLVR methods primarily rely on sparse, binary feedback from verifiers, which lacks the resolution to distinguish the quality of intermediate reasoning steps.




\vspace{-1mm}
\subsection{Contrastive Learning}
Contrastive learning (CL) has emerged as a powerful paradigm for self-supervised representation learning, aiming to map data points into a latent space where semantically similar samples are clustered together while dissimilar ones are pushed apart~\citep{oord2018representation,frosst2019analyzing,khosla2020supervised,chen2020simple}. 
A cornerstone of CL methods is the InfoNCE objective~\citep{oord2018representation}:
Formally, given a group of pairwise instances $\{(\vx_i,\vy_i)\}_{i}^N$ , CL seeks to optimize the embedding model as a representation function $f(\cdot)$ such that it minimizes the distance between each $\vx_i$ and its positive counterpart $\vy_i$, while simultaneously maximizing $\vx_i$'s  distance from a set of negative samples $\{\vy_j\}_{j\neq i}$.
A widely adopted approach to realize this objective is the InfoNCE loss~\citep{oord2018representation}, which frames the representation learning task as a multi-class categorical classification problem:
\begin{equation}
      \gL_{\text{InfoNCE}}:= - \bbE_{p(\vx,\vy)}\Big[\frac{1}{N} \sum_{i = 1}^N \log \frac{\exp(f(\vx_i,\vy_i))}{\sum_{j=1}^N \exp(f(\vx_i, \vy_j))} \Big], \label{eq:NCE}
\end{equation}
which frames the learning process as a multi-class categorical cross-entropy task to identify a positive sample among a set of noise contrastive samples. 

Theoretically, the negative InfoNCE loss is equivalent to a lower-bound estimator of the mutual information (MI) $\MI(\vx;\vy)$ between the two variables $(\vx,\vy)$ with samples $\{(\vx_i,\vy_i) \}_{i=1}^N$ \citep{poole2019variational}:
\begin{equation}\label{eq:mutual_info_def}
    \MI(\vx;\vy): = \bbE_{p(\vx,\vy)}\Big[\log \frac{p(\vx,\vy)}{p(\vx)p(\vy)}\Big] \geq \log(N) -\gL_\text{InfoNCE},
\end{equation}
where $p(\vx,\vy)$ is the joint distribution of $(\vx,\vy)$, and $p(\vx)$ and $p(\vy)$ are the marginal distribution of $\vx$ and $\vy$ respectively.
Hence, by minimizing the InfoNCE loss, one can enlarge the information overlap between the two learnable features or embeddings, thereby distilling the underlying semantic invariants. Specifically, the loss encourages the representations of positive pairs to be close in the embedding space while pushing negative pairs apart, effectively capturing the shared features across similar instances.


\begin{figure*}[t]
    \centering
    \includegraphics[width=0.9\textwidth]{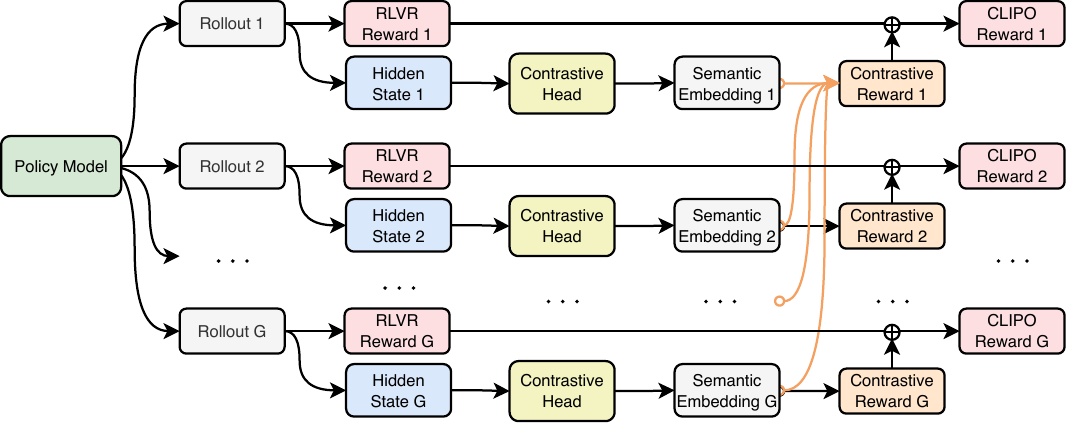}
    \caption{Framework of \ours. For each input prompt $\vx$, policy optimization methods generate a group of rollouts $\{\vy_1,\vy_2,\dots,\vy_G\}$ then calculate corresponding RLVR rewards $\{r_1,r_2,\dots,r_G\}$.  CLIPO applies a contrastive head on the top of the last hidden states $\{\vh_1,\vh_2,\dots,\vh_G\}$ of the rollout group and outputs trajectory-level semantic embeddings $\{\ve_1,\ve_2,\dots,\ve_G\}$. The contrastive rewards $\{r_1^\text{CL},r_2^\text{CL},\dots,r_G^\text{CL}\}$ are computed across the semantic embedding group to provide the similarity of successful and failed trajectories. The final CLIPO reward for the $i$-th rollout is $r'_i=r_i+ r^\text{CL}_i$.    
    }
    \label{fig:framework}
\end{figure*}

\vspace{-1mm}
\section{Methodology}
\vspace{-1mm}
%

We propose \textbf{C}ontrastive \textbf{L}earning \textbf{i}n \textbf{P}olicy \textbf{O}ptimization (CLIPO), a framework designed to enhance the reasoning capabilities of LLMs by exploiting the latent semantic consistency among successful reasoning trajectories. As illustrated in Figure \ref{fig:framework}, CLIPO integrates an intra-group contrastive reward into standard policy optimization. Instead of relying solely on sparse, outcome-based feedback, we augment the reward signal with a dense contrastive component derived from a latent embedding space that promotes alignment among correct reasoning paths and encourages separation from erroneous ones. Specifically, we append a lightweight contrastive head to the LLM backbone, which extracts trajectory-level representations from the hidden states. During training, we compute a contrastive loss within each rollout group, using correct responses as positive anchors and incorrect ones as negatives. This loss is subsequently repurposed as a dense auxiliary reward, providing the model with a more informative signal to guide policy updates. We formalize the CLIPO objective in Section \ref{sec:method_objective} and detail our implementation strategy in Section \ref{sec:method_implementation}.

\subsection{\ours Objective}
\label{sec:method_objective}

Suppose we have a binary reward function $r:\gX \times \gY \rightarrow \{0,1\}$ where $r(\vx,\vy) = 1$ if the model response $\vy$ is correct, and $r(\vx,\vy) = 0$ if the model response is incorrect. The mainstream reinforcement learning with variable rewards (RLVR) aims to optimize the policy optimization objective in \eqref{eq:rlvr_objective}.
%
As discussed in Figure~\ref{fig:intro}, we expect the policy model $\pi_\theta(\vy|\vx)$ to identify the commonalities among the correct rollout samples during the RLVR process, \textit{i.e.}, to maximize the mutual information (as in \eqref{eq:mutual_info_def}) among the positive rollouts:
%
\begin{equation} \label{eq:mutual_info_between_pos_pair}
    \max_{\vy,\bar{\vy} \sim \pi_\theta(\cdot|\vx)} \MI\Big(\vy; \bar{\vy} \big\vert r(\vx,\vy) = 1, r(\vx,\bar{\vy}) =1\Big).
\end{equation}
Note that the mutual information calculation is conditioned on the event $\gP_{\vy,\bar{\vy}}:=\{ r(\vx,\vy)=1, r(\vx,\bar{\vy}) = 1 \}$, where $\vy$ and $\bar{\vy}$ correctly obtain positive rewards. By maximizing \eqref{eq:mutual_info_between_pos_pair}, the policy model $\pi_\theta$ is required to seek the commonalities among positive rollouts, obtaining more robust experiences of successes for policy optimization. 
Therefore, the overall learning objective for CLIPO is:
\begin{equation}\label{eq:CLIPO_obj}
    \max_{\vy,\bar{\vy} \sim \pi_\theta(\cdot|\vx)}  \bbE\big[r(\vx,\vy)\big] + \lambda \cdot \MI\big(\vy;\bar{\vy} \big\vert \vx, \gP_{\vy,\bar{\vy}}\big). 
\end{equation}
By maximizing the mutual information among positive rollouts, we enforce semantic proximity among successful reasoning paths. This regularization provides a dense, informative signal that guides the policy toward logically coherent reasoning, effectively compensating for the limitations of sparse outcome-based verifiers of RLVR.

Pracitically, the exact MI value $\MI\big(\vy;\bar{\vy} \big\vert \vx, \gP_{\vy,\bar{\vy}}\big)$ is challenging to compute, for the joint distribution $p(\vy,\bar{\vy} | \vx, \gP_{\vy,\bar{\vy}})$ is intractable~\citep{cheng2020club}. Fortunately, in each policy optimization step, we obtain a group of rollouts $\gG=\{\vy_1, \vy_2, \dots, \vy_G\} \sim \pi_\theta(\cdot|\vx)$, where we can use a sample-based MI estimator to approximate the ground-truth value of $\MI\big(\vy;\bar{\vy} \big\vert \vx, \gP_{\vy,\bar{\vy}}\big)$. More specifically, we select the InfoNCE~\citep{oord2018representation} as the lower bound approximation for MI maximization, \textit{i.e.}, for each $\vx$, $\MI\big(\vy;\bar{\vy} \big\vert \vx, \gP_{\vy,\bar{\vy}}\big) \geq$
\begin{equation}\label{eq:contrastive_infoNCE}
    \log(G) + \frac{1}{|\gP|} \sum_{\vy, \bar{\vy}\in \gP} \log \left(\frac{\exp(f(\vy,\bar{\vy}))}{\sum_{i=1}^G \exp(f(\vy,\vy_i))}\right) = \log(G) - \gL_\text{CL}(\vx),
\end{equation}
where $\gP = \{\vy_i \in \gG: r(\vx,\vy_i)=1\}$ denotes the rollout subset with successful outcome. For each positive rollout $\vy$, we uniformly sample another positive rollout $\bar{\vy}\in\gP$ to make up the \textit{positive} pair $(\vy,\bar{\vy})$, and use the remaining rollouts in group as the \textit{negatives} $\{(\vy,\vy_i)\} \mid \vy_i \in \gG, \vy_i \neq \bar{\vy}\}$. Then the InfoNCE loss is computed as the log-ratio of the positive pair score to the mean score of negative pairs in a contrastive learning scheme. The score function $f(\vy,\bar{\vy})$ measures the semantic similarity between $\vy$ and $\bar{\vy}$, which we implement as the inner product of embeddings, following the design from SimCLR~\citep{chen2020simple}:
\begin{equation}
f(\vy,\bar{\vy}) = g_\phi(\vh_\theta(\vy))^\mathrm{T} g_\phi(\vh_\theta(\bar{\vy}))/ \tau,
\end{equation}
where $\vh_\theta(\vy)$ is the last hidden state output by the policy model $\pi_\theta(\cdot|\vx)$'s transformer backbone, and $g_\phi(\cdot)$ is a learnable contrastive head to extract core semantic information from the last hidden state of the transformer. $\tau > 0$ is the temperature hyper-parameter. 
Note that the contrastive loss in \eqref{eq:contrastive_infoNCE} is inapplicable when $|\gP| \leq 1$ (unable to provide positive pairs) or $|\gP|=G$ (no failures). Therefore, we only apply the CLIPO loss on samples satisfying $1< |\gP|<G$ to avoid these degenerate cases.

\subsection{\ours Implementation}
\label{sec:method_implementation}

\paragraph{Sentence-Level Representation.}
For each response $\vy$, the policy model $\pi_\theta(\cdot|\vx)$ produces token-level hidden states $\bar{\vh}_\theta(\vy) \in \mathbb{R}^{T \times D}$, where $T$ is the length of the response token sequence and $D$ is the dimension of hidden states.
To derive a sentence-level representation from the token-level hidden states, we apply a mean pooling operation of all response hidden states across the sequence dimension:
\vskip -0.1in
\begin{equation}
    {\vh}_\theta(\vy) = \frac{1}{T} \sum_{t=1}^T \bar{\vh}_\theta(\vy)_t,
\end{equation}
obtaining ${\vh}_\theta(\vy) \in \mathbb{R}^{D}$ that encapsulates the overall semantic content of the response.
This pooled representation is then passed through the contrastive head $g_\phi: \mathbb{R}^{D} \rightarrow \mathbb{R}^{d}$ to obtain the semantic embedding:
$\ve(\vy) = g_\phi({\vh}_\theta(\vy)),$
where $\ve(\vy) \in \mathbb{R}^{d}$ is used for similarity calculation in contrastive loss computation.

\paragraph{Contrastive Head Optimization.} 
The contrastive head $g_\phi$ typically is a simple linear layer $W \in \mathbb{R}^{d \times D}$, designed to project the pooled representation into a latent space suitable for contrastive learning~\citep{chen2020simple}.
Then the contrastive loss for each correct response $\vy_i$ in the group $\gG$ is:
\begin{equation}\label{eq:contrastive_loss_sample}
    \mathcal{L}_\text{CL}(\vx,\vy_i) = - \log \left(\frac{\exp(f(\vy_i,\bar{\vy}_i))}{\sum_{j=1}^G \exp(f(\vy_i,\vy_j))}\right),
\end{equation}
where $\bar{\vy}_i$ is a uniformly sampled positive sample from the set of other correct responses in the group $\gG$.
The contrastive head is updated jointly with the policy model during training, and its parameters $\phi$ are optimized to minimize the average contrastive loss across all valid anchors in the batch.
The anchor with no positive pair in the group is considered invalid and excluded from loss computation.

\paragraph{Contrastive Reward Integration.}
The total reward $r'_i$ for a given response $y_i$ is formulated as the sum of the original verifiable reward $r_i$ and the contrastive reward $r_{i}^{\text{CL}}$:
\begin{equation}
    r_{i}' = r_{i} + r_{i}^{\text{CL}},
\end{equation}
To prevent the auxiliary signal from dominating the verifiable objective, we scale it by the re-weighting parameter $\lambda>0$ in \eqref{eq:CLIPO_obj} and clip it with a lower threshold:
\begin{equation}
    r_{i}^{\text{CL}} = \max(-\lambda \cdot \mathcal{L}_\text{CL}(\vx,\vy_i), -0.5),
\end{equation}
where $\mathcal{L}_\text{CL}(\vx,\vy_i)$ is the contrastive loss in \eqref{eq:contrastive_loss_sample}. 
%
By incorporating $r_{i}^{\text{CL}}$, we move beyond the limitations of sparse, binary feedback of RLVR. This mechanism provides granular guidance even among multiple correct responses by prioritizing those that converge within the ``consensus'' region of the solution space. Consequently, the model is compelled to refine its reasoning strategies toward more consistent and robust trajectories, rather than merely satisfying the verification threshold.

\vspace{-1mm}
\section{Experiments}
\vspace{-1mm}
\subsection{Experimental Setup}

\paragraph{Datasets.} To rigorously evaluate the effectiveness of our proposed method, we construct two separate experimental tracks using training sets of varying difficulty levels.
\begin{itemize}[leftmargin=*]
\vspace{-2mm}
\item \textbf{Track I: Grade-School and General Reasoning.} 
In the first setup, we fine-tune our model on the GSM8K dataset, which comprises approximately 8,000 high-school-level mathematical word problems. 
We evaluate it on the GSM8K test set, as well as mathematical benchmarks including GSM8K-Symbolic, GSM8K-P1, GSM8K-P2~\citep{mirzadeh2025gsmsymbolic}. 
These variants introduce different levels of complexity and distribution shifts to assess the model's generalization capabilities.
Furthermore, we assess its broader reasoning and general knowledge capabilities across diverse benchmarks such as CommonsenseQA~\citep{talmor2019commonsenseqa}, TruthfulQA~\citep{lin2022truthfulqa}, TheoremQA~\citep{chen2023theoremqa}, and the MMLU~\citep{hendrycks2021mmlu} suite.
We use SYM, P1, P2, cQA, tQA, thrmQA to denote GSM8K-Symbolic, GSM8K-P1, GSM8K-P2, CommonsenseQA, TruthfulQA, TheoremQA datasets respectively.

\item \textbf{Track II: Competition-Level Mathematical Reasoning.} 
To further investigate the impact of task difficulty on model performance, we conduct an additional set of experiments by training on the MATH 7.5k dataset~\citep{hendrycks2021measuring}. 
The evaluation datasets include MATH500~\citep{hendrycks2021measuring}, Math-Perturb~\citep{huang2025mathperturb} (covering simple and hard variants), which further test robustness to problem perturbations.
Meanwhile, we also assess performance on competition-level challenges such as AMC23, AIME and AIME25, which feature problems of significantly higher complexity.
We use Math, Math\_S, Math\_H to denote MATH500, Math-Perturb Simple, Math-Perturb Hard datasets respectively.

\end{itemize}
\paragraph{Metrics.} 
We report Pass@1~\citep{dai2025sgrpo} as our primary evaluation metric. In Track I, we generate a single response using a temperature of 0.6. In Track II, given the higher difficulty and limited size of the datasets, we sample 16 responses per question with the same temperature of 0.6.

\paragraph{Models.}
For experiments on Track I, we utilize Qwen2.5-3B-Instruct as the base model. 
For the more challenging MATH7.5k dataset, we evaluate our method across Qwen2.5-7B-Instruct, Llama3.1-8B-Instruct, and reasoning DeepSeek-R1-Distill-Qwen-7B.

\paragraph{Baselines.}
We compare \ours against several state-of-the-art group-based RLVR baselines, including: 
\begin{itemize}[leftmargin=*, itemsep=1pt]
\vspace{-2mm}
    \item \textbf{Base Model}: The original models without any additional training. 
    \item \textbf{GRPO}~\citep{shao2024deepseekmath}: An efficient and effective algorithm that eliminates the need for a separate critic model. It estimates the baseline by averaging rewards from a group of responses for the same prompt.
    \item \textbf{GSPO}~\citep{zheng2025gspo}: An extension of group-based optimization that shifts focus from token-level to sequence-level likelihood and clipping.
    \item \textbf{DAPO}~\citep{yu2025dapo}: It introduces techniques such as decoupled clipping ranges and dynamic sampling.
    \item \textbf{GMPO}~\citep{zhao2025gmpo}: A variant of RLVR that maximizes the geometric mean of reward-weighted policy ratios instead of the arithmetic mean.

\end{itemize}

\paragraph{Contrastive Head Designs.}

We implement the contrastive head using a Linear layer, following designs from prior work~\citep{chen2020simple,khosla2020supervised}.
The input dimension $D$ of the contrastive head matches the hidden size of the respective base model. 
For track I, the output dimensionality $d=512$. For track II, we increase the output size to $2048$ to provide the higher representational capacity required for complex mathematical reasoning steps. 
The final embeddings are obtained via $L_2$-normalization. 
The head is optimized using AdamW with a learning rate of $1 \times 10^{-3}$ and a weight decay of $0.01$. 
We experiment with various contrastive loss functions, including InfoNCE loss~\citep{oord2018representation,chen2020simple}, SupCon loss~\citep{khosla2020supervised} and Soft Nearest Neighbor loss~\citep{salakhutdinov2007learning,frosst2019analyzing}.
In all configurations, we use the coefficient $\lambda=0.2$ for InfoNCE and SupCon, and $\lambda=1$ for SoftNN loss.

\paragraph{Implementation.}
Our training pipeline is implemented using the VERL framework~\citep{sheng2024hybridflow}.
For track I, we use a global batch size of 512. The models are trained for 5 epochs, before which we perform a 1-epoch warmup for the contrastive head. 
The maximum response length is restricted to 2048 tokens, with default 16 rollouts per prompt. 
For track II, we use a batch size of 128, extending the training to 8 epochs, with a similar early warmup for the head.
To accommodate the complexity of the problems, the maximum response length is increased to 4096 tokens. 
For most mathematical reasoning datasets (e.g., GSM8K, MATH, AMC, AIME, AIME2025, TheoremQA, and their variants), we adopt the following instruction:
\noindent\texttt{Let's think step by step and output the final answer within \textbackslash boxed\{\}.}
For multiple-choice datasets (e.g., MMLU, TruthfulQA, CommonsenseQA), we further specify the expected answer format by including the available options.
For example, for MMLU:
\noindent\texttt{Let's think step by step and output the final answer (e.g., A, B, C, D) within \textbackslash boxed\{\}.}
These setting is consistent across all baselines to ensure fair comparisons.
Appendix~\ref{sec:app_implementation_details} provides further details.

\begin{table*}[h]
\centering
\caption{Performance comparison of different RLVR methods on Track I(GSM8K and General Reasoning). 
`Avg.' represents the average performance across all datasets, and `M-Avg.' represents the average performance on math-related datasets. `G-Avg.' represents the average performance on general and QA tasks.
`$\Delta$' represents the improvement after using our method, with light green indicating improvement and light red indicating decline.
The best results are highlighted in \textbf{bold}, and the second-best results are \underline{underlined}.
}
\label{tab:main_results_gsm8k}
\resizebox{\textwidth}{!}{%
\begin{tabular}{l *{11}{S[table-format=2.2, table-column-width=0.98cm]}}
\toprule
\textbf{Methods} & \textbf{GSM8K} & \textbf{SYM} & \textbf{P1} & \textbf{P2} & \textbf{cQA} & \textbf{tQA} & \textbf{thrmQA} & \textbf{MMLU} & \textbf{M-Avg.} & \textbf{G-Avg.} & \textbf{Avg.} \\
\midrule
Base Model& 86.05 & 82.40 & 70.90 & 47.96 & 73.63 & 29.99 & 24.90 & 68.09 & 71.83 & 49.15 & 60.49 \\
GRPO & 87.79 & 84.04 & 73.12 & 50.80 & 74.94 & 30.48 & \textbf{27.71} & 68.19 & 73.94 & 50.33 & 62.13 \\
GSPO & 87.19 & 83.94 & 73.28 & 51.36 & 74.86 & 30.23 & 27.38 & 67.68 & 73.94 & 50.04 & 61.99 \\
DAPO & 87.41 & 84.56 & \underline{74.70} & 52.32 & 76.00 & 31.58 & 27.11 & \textbf{69.00} & 74.75 & 50.92 & 62.84 \\
GMPO & 86.81 & 84.02 & 73.90 & 49.80 & 76.33 & 30.97 & 26.57 & 68.49 & 73.63 & 50.59 & 62.11 \\
\hline
GRPO+\ours & \textbf{88.02} & 84.62 & 74.60 & \textbf{54.16} & \underline{76.90} & 31.52 & \underline{27.58} & \underline{68.64} & \textbf{75.35} & \textbf{51.16} & \textbf{63.26} \\
\cellcolor{gray!10}$\Delta$ & \cellcolor{pos}0.23 & \cellcolor{pos}0.58 & \cellcolor{pos}1.48 & \cellcolor{pos}3.36 & \cellcolor{pos}1.97 & \cellcolor{pos}1.04 & \cellcolor{neg}-0.13 & \cellcolor{pos}0.46 & \cellcolor{pos}1.41 & \cellcolor{pos}0.83 & \cellcolor{pos}1.12 \\
GSPO+\ours & \underline{87.95} & 83.98 & 74.54 & 52.92 & 75.92 & \textbf{32.19} & 26.10 & 67.48 & 74.85 & 50.42 & 62.63 \\
\cellcolor{gray!10}$\Delta$ & \cellcolor{pos}0.76 & \cellcolor{pos}0.04 & \cellcolor{pos}1.26 & \cellcolor{pos}1.56 & \cellcolor{pos}1.06 & \cellcolor{pos}1.96 & \cellcolor{neg}-1.27 & \cellcolor{neg}-0.20 & \cellcolor{pos}0.90 & \cellcolor{pos}0.39 & \cellcolor{pos}0.65 \\
DAPO+\ours & 87.57 & \underline{84.66} & 74.54 & \underline{53.12} & \textbf{77.07} & 31.40 & 27.44 & 67.93 & \underline{74.97} & \underline{50.96} & \underline{62.97} \\
\cellcolor{gray!10}$\Delta$ & \cellcolor{pos}0.15 & \cellcolor{pos}0.10 & \cellcolor{neg}-0.16 & \cellcolor{pos}0.80 & \cellcolor{pos}1.06 & \cellcolor{neg}-0.18 & \cellcolor{pos}0.33 & \cellcolor{neg}-1.06 & \cellcolor{pos}0.22 & \cellcolor{pos}0.04 & \cellcolor{pos}0.13 \\
GMPO+\ours & 87.64 & \textbf{85.52} & \textbf{75.52} & 50.92 & 75.35 & \underline{31.95} & \underline{27.58} & 67.63 & 74.90 & 50.63 & 62.76 \\
\cellcolor{gray!10}$\Delta$ & \cellcolor{pos}0.83 & \cellcolor{pos}1.50 & \cellcolor{pos}1.62 & \cellcolor{pos}1.12 & \cellcolor{neg}-0.98 & \cellcolor{pos}0.98 & \cellcolor{pos}1.00 & \cellcolor{neg}-0.86 & \cellcolor{pos}1.27 & \cellcolor{pos}0.03 & \cellcolor{pos}0.65 \\
\bottomrule
\end{tabular}
}
\end{table*}

\subsection{Main Results}

Tables \ref{tab:main_results_gsm8k} and \ref{tab:main_results_math} summarize the performance of our method compared to several state-of-the-art RLVR baselines.
Since \ours is compatible with group-based methods, we denote `method+\ours' or `\ours(method)' as the version of method augmented with \ours. We use GRPO as the default base method, and $\Delta$ indicates the relative improvement brought by \ours.

\paragraph{Results on GSM8K and General Reasoning.}
As shown in Table \ref{tab:main_results_gsm8k}, GRPO+\ours achieves the highest overall average score of 63.26, consistently outperforming all baselines.
Our method shows significant improvements on the more challenging variants of GSM8K. Specifically, GRPO+\ours improves GSM8K-P1 and GSM8K-P2 by $+1.48$ and $+3.36$ points, respectively, and achieves the highest math-average score of $75.35$. 
These results indicate that contrastive reward signals are particularly beneficial for enhancing robustness under distribution shifts and more challenging compositional reasoning settings.
Beyond mathematics, \ours exhibits strong cross-domain generalization.
\ours attains either the best or second-best performance across all four general reasoning benchmarks. 
This indicates that the contrastive auxiliary reward enhances the model's general reasoning capabilities without sacrificing its general knowledge or linguistic capabilities.

\begin{table*}[t]
\centering
\caption{Performance comparison on Track II(Competition-Level Reasoning). 
`M-Avg.' represents the average performance on Math, Math\_S, Math\_H. `C-Avg.' represents the average performance on competition-level datasets including AMC, AIME, and AIME25.
}
\label{tab:main_results_math}
\resizebox{\textwidth}{!}{%
\begin{tabular}{l *{9}{S[table-format=2.2, table-column-width=1.2cm]}}
\toprule
\textbf{Methods} & \textbf{Math} & \textbf{Math\_S} & \textbf{Math\_H} & \textbf{AMC} & \textbf{AIME} & \textbf{AIME25} & \textbf{M-Avg.} & \textbf{C-Avg.} & \textbf{Avg.} \\
\midrule
Base Model& 75.35 & 59.24 & 35.31 & 44.13 & 10.63 & 6.25 & 56.63 & 20.33 & 38.48 \\
GRPO & 76.46 & 64.83 & 37.86 & 46.84 & 17.92 & 9.58 & 59.72 & 24.78 & 42.25 \\
GSPO & \underline{77.58} & 64.56 & 38.36 & \underline{50.08} & 16.25 & 10.42 & 60.17 & 25.58 & 42.87 \\
DAPO & 76.03 & 66.08 & \textbf{39.94} & 47.36 & 19.79 & 7.92 & 60.68 & 25.02 & 42.85 \\
GMPO & \textbf{78.47} & 64.27 & 37.25 & 48.42 & 19.58 & 9.58 & 60.00 & 25.86 & 42.93 \\
\hline
GRPO+\ours & 77.49 & \textbf{67.21} & 38.97 & 49.17 & 18.13 & 10.63 & 61.22 & 25.97 & 43.60 \\
\cellcolor{gray!10}$\Delta$ &
\cellcolor{pos}1.03 &
\cellcolor{pos}2.38 &
\cellcolor{pos}1.11 &
\cellcolor{pos}2.33 &
\cellcolor{pos}0.21 &
\cellcolor{pos}1.04 &
\cellcolor{pos}1.50 &
\cellcolor{pos}1.19 &
\cellcolor{pos}1.35 \\
GSPO+\ours & 77.16 & 66.69 & 37.66 & \textbf{50.53} & \textbf{20.63} & 9.38 & 60.50 & \textbf{26.84} & 43.67 \\
\cellcolor{gray!10}$\Delta$ &
\cellcolor{neg}-0.41 &
\cellcolor{pos}2.13 &
\cellcolor{neg}-0.70 &
\cellcolor{pos}0.45 &
\cellcolor{pos}4.38 &
\cellcolor{neg}-1.04 &
\cellcolor{pos}0.34 &
\cellcolor{pos}1.26 &
\cellcolor{pos}0.80 \\
DAPO+\ours & 77.43 & 66.55 & \underline{39.82} & 48.42 & 18.75 & \textbf{13.33} & \underline{61.27} & \underline{26.83} & \textbf{44.05} \\
\cellcolor{gray!10}$\Delta$ &
\cellcolor{pos}1.40 &
\cellcolor{pos}0.48 &
\cellcolor{neg}-0.11 &
\cellcolor{pos}1.05 &
\cellcolor{neg}-1.04 &
\cellcolor{pos}5.42 &
\cellcolor{pos}0.59 &
\cellcolor{pos}1.81 &
\cellcolor{pos}1.20 \\
GMPO+\ours & 77.87 & \underline{67.14} & 38.83 & 47.67 & \underline{19.38} & \underline{11.67} & \textbf{61.28} & 26.24 & \underline{43.76} \\
\cellcolor{gray!10}$\Delta$ &
\cellcolor{neg}-0.60 &
\cellcolor{pos}2.88 &
\cellcolor{pos}1.58 &
\cellcolor{neg}-0.75 &
\cellcolor{neg}-0.21 &
\cellcolor{pos}2.08 &
\cellcolor{pos}1.28 &
\cellcolor{pos}0.37 &
\cellcolor{pos}0.83 \\
\bottomrule
\end{tabular}
}
\end{table*}

\paragraph{Results on Competition-Level Reasoning.}
Table~\ref{tab:main_results_math} presents the results for competition-level mathematical reasoning tasks.
DAPO+\ours achieves the highest average score of 44.05, surpassing all baselines. GMPO+\ours attains the second-best score of 43.76.
Specifically, integrating \ours leads to performance gains across GRPO, GSPO, DAPO, and GMPO in overall average scores by $+1.35$, $+0.80$, $+1.20$, and $+0.83$ respectively. 
On specific datasets such as MATH500, Math-Perturb Simple, and Math-Perturb Hard, \ours yields average improvements of $+1.50$, $+0.34$, $+0.59$, and $+1.28$ respectively, 
demonstrating the effectiveness in enhancing reasoning capabilities under perturbations and distribution shifts.
This highlights the efficacy of the contrastive head in improving the model's generalization abilities.
On competition-level mathematical datasets, \ours achieves average improvements of $+1.19$, $+1.26$, $+1.81$, and $+0.37$ across the four different RLVR methods, respectively. 
This demonstrates the effectiveness of \ours in high-difficulty reasoning tasks.
Overall, all of \ours outperform their respective baselines, demonstrating the universality of the contrastive reward mechanism in enhancing reasoning capabilities and generalizability.

Across both tracks, the gains of \ours are consistent, especially on more challenging or distribution-shifted mathematical benchmarks. 
In out-of-distribution settings such as perturbation and symbolic reasoning tasks, \ours demonstrates enhanced robustness and generalization capabilities.
These results validate our hypothesis that contrastive rewards, which encourage clustering of high-quality solutions in the representation space while pushing away low-quality ones, provide more informative and fine-grained learning signals than binary outcome-based feedback alone. 
Importantly, this enhancement is largely orthogonal to the underlying RLVR algorithm, offering complementary signals that capture fine-grained distinctions among trajectories.


\subsection{Ablation Studies and Analysis}

\begin{table}[b]
\centering
\caption{Comparison between \ours with and without fixed head on Track I and Track II. Metric 1 and Metric 2 represent M-Avg., G-Avg. for Track I, and M-Avg., C-Avg. for Track II, respectively. $\Delta$ indicates the performance drop when using fixed head.
}
\label{tab:more_fixedhead}
\begin{tabularx}{0.8\linewidth}{XXXXX}
\toprule
\textbf{Track} & \textbf{Methods} & \textbf{Metric 1} & \textbf{Metric 2} & \textbf{Avg.} \\
\midrule
\multirow{3}{*}{Track I} 
& \ours        & 75.35 & 51.16 & 63.26 \\
& \ours-fixed  & 74.73 & 50.24 & 62.48 \\
& \cellcolor{gray!10}$\Delta$     & \cellcolor{neg}-0.62 & \cellcolor{neg}-0.92 & \cellcolor{neg}-0.77 \\
\midrule
\multirow{3}{*}{Track II}
& \ours        & 61.22 & 25.97 & 43.60 \\
& \ours-fixed  & 60.19 & 25.06 & 42.63 \\
& \cellcolor{gray!10}$\Delta$     & \cellcolor{neg}-1.03 & \cellcolor{neg}-0.91 & \cellcolor{neg}-0.97 \\
\bottomrule
\end{tabularx}
\end{table}


\paragraph{Contrastive Head Analysis.}
\ours introduces a contrastive head that maps the LLM's hidden states into a new embedding space, thereby learning a novel geometric structure.
To validate the effectiveness of the contrastive head, we conduct an ablation where the contrastive head is fixed and not updated during training (denoted as \ours-fixed). 
As shown in Table~\ref{tab:more_fixedhead}, freezing the head leads to consistent performance drops across all benchmarks in both tracks. 
In Track I, \ours-fixed results in decreases of $-0.62$, $-0.92$, and $-0.77$ in math-average, general-average, and overall-average scores, respectively.
In Track II, \ours-fixed yields reductions of $-1.03$, $-0.91$, and $-0.97$ in there scores, respectively.
These results highlight the importance of jointly optimizing the contrastive head with the base model.
By allowing the head to adapt during training, the model can learn more effective representations that better capture the nuances of high-quality versus low-quality solutions, thereby enhancing the overall learning process.
The complete experimental results, along with a visualization analysis of the semantic embeddings, are presented in Appendix~\ref{sec:app_contrastive_head}.




\begin{figure}[h]
\centering

\begin{minipage}[h]{0.48\linewidth}
\centering
\captionof{table}{Comparison among contrastive loss variants. 
Here, $s_{ia}$ denotes the similarity between $i$ and $a$, while $P$ is the set of positive examples for $i$ and $p^*$ is a single sampled positive example.
}
\label{tab:more_contrastiveloss}

\resizebox{\linewidth}{!}{%
\begin{tabular}{lcc}
\toprule
\textbf{Loss} & \textbf{Formulation} & \textbf{Positive Aggregation} \\
\midrule
InfoNCE & $-\log \frac{e^{s_{ip^*}}}{\sum_a e^{s_{ia}}}$ & Single Positive \\
SupCon & $-\frac{1}{|P|}\sum_p \log \frac{e^{s_{ip}}}{\sum_a e^{s_{ia}}}$ & Mean Positives \\
SoftNN & $-\log \frac{\sum_p e^{s_{ip}}}{\sum_a e^{s_{ia}}}$ & Sum Positives \\
\bottomrule
\end{tabular}
}
\end{minipage}
\hfill
\begin{minipage}[h]{0.48\linewidth}
\centering
\includegraphics[width=\linewidth]{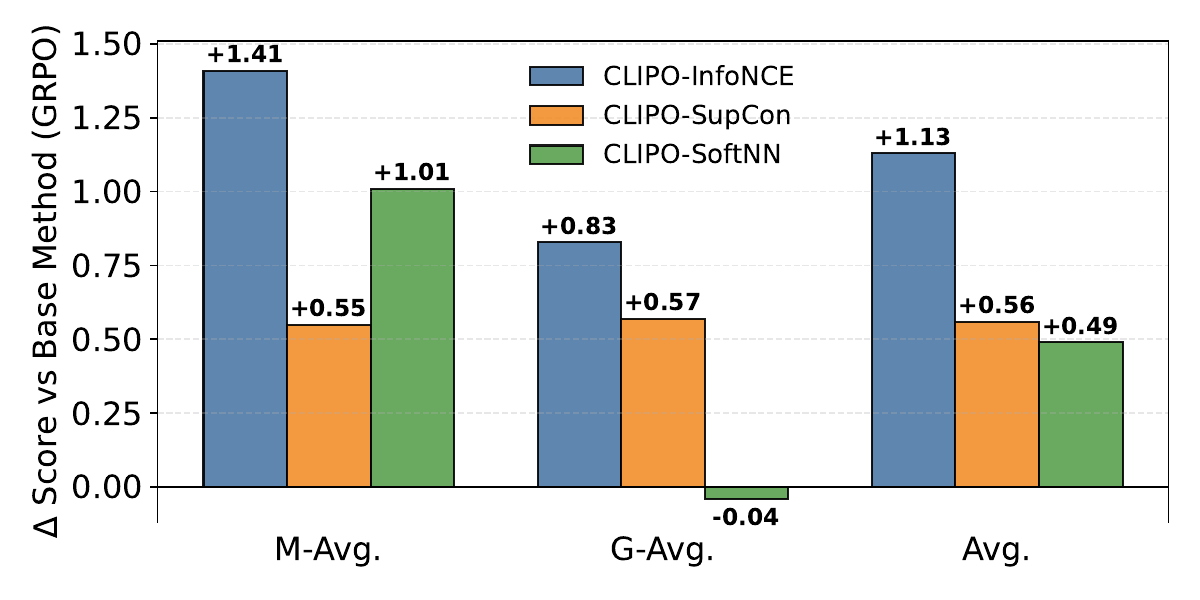}
\captionof{figure}{Performance Gain Across Different Losses.}
\label{fig:more_contrastiveloss}
\end{minipage}

\end{figure}


\paragraph{Contrastive Loss Variants.}
Except for the InfoNCE loss, we also experiment with two contrastive loss functions, SupCon loss~\citep{khosla2020supervised} and SoftNN loss~\citep{salakhutdinov2007learning,frosst2019analyzing}. 
The main differences among these loss functions are summarized in Table~\ref{tab:more_contrastiveloss}.
Specifically, the difference lies in the strategy for selecting positives. InfoNCE samples a single correct response as the positive, SupCon loss considers all correct responses as positives and computes the average outside the logarithm, whereas SoftNN loss also uses all correct responses but computes their sum inside the logarithm.
As shown in Figure~\ref{fig:more_contrastiveloss}, all three contrastive loss variants yield notable performance improvements over the base GRPO method on Track I.
InfoNCE loss achieves the highest gains(+1.13) across three average scores, followed closely by SupCon loss(+0.56).
SoftNN shows insufficiently stable improvements;
however, it still exceeds the baseline by +0.49 on the overall average.
More detailed results, as well as experiments on Track II, can be found in Appendix~\ref{sec:app_more_losses}.
These results suggest that while the specific choice of contrastive loss can influence performance, the overall benefit of incorporating contrastive rewards is robust across different formulations.

\begin{wraptable}{r}{0.5\textwidth}
\centering
\caption{Effect of different contrastive learning temperature $\tau$ on \ours performance. 
It can be observed that lower temperature coefficients lead to better performance improvements; however, excessively low temperature may result in degraded performance.
}
\label{tab:more_temperature}

\resizebox{0.9\linewidth}{!}{%
\begin{tabular}{lccc}
\toprule
\textbf{$\tau$} & \textbf{Math-Avg.} & \textbf{Comp-Avg.} & \textbf{Avg.} \\
\midrule
$0.2$   & 59.90 & 24.36 & 42.13 \\
$0.1$   & 59.76 & \textbf{26.81} & \underline{43.28} \\
$0.05$  & \textbf{61.28} & 24.85 & 43.07 \\
$0.02$  & \underline{61.22} & \underline{25.97} & \textbf{43.60} \\
$0.015$ & 60.64 & 25.19 & 42.91 \\
\bottomrule
\end{tabular}
}

\end{wraptable}

\paragraph{Contrastive Temperature $\tau$ Analysis.}
The temperature parameter $\tau$ plays a crucial role in contrastive learning. 
We conduct experiments to evaluate the impact of varying temperature values on training effectiveness. The results are presented in Table~\ref{tab:more_temperature}.
From the table, it is evident that lower temperature values generally yield better performance improvements. 
Specifically, reducing $\tau$ from 0.2 to 0.02 improves the overall average score from 42.13 to 43.60, indicating that sharper similarity scaling benefits representation learning in our setting.
This is attributed to the model's increased focus on hard negatives, which enhances its discriminative capabilities, aligning with findings from~\citet{wang2020understanding,wang2021understanding}.
Meanwhile, higher temperature settings (e.g., 0.2) lead to performance degradation.
To analyze this phenomenon, we plot the average cosine similarity of positive pairs within the same group during training, as shown in Appendix~\ref{sec:app_temperature}.
We observe that higher temperature in greater fluctuations in the similarity of positive pairs, indicating that the model struggles to consistently push positive pairs together during training.
Our findings suggest that lower temperatures are more effective for contrastive learning in our context. 
One possible explanation is that our setting involves a significant number of hard negatives, and a lower temperature enhances the model's discriminative power, leading to better training outcomes~\citep{wang2020understanding}. 


\begin{table*}[t]
\centering
\caption{Effect of different group sizes on \ours performance. Larger group sizes lead to better performance improvements.
The best results are highlighted in bold. It can be observed that larger group sizes generally lead to better performance improvements across Avg. metrics.
}
\label{tab:more_group_size}
\resizebox{\textwidth}{!}{%
\begin{tabular}{l *{9}{S[table-format=2.2, table-column-width=1.2cm]}}
\toprule
\textbf{GroupSize} & \textbf{Math} & \textbf{Math\_S} & \textbf{Math\_H} & \textbf{AMC} & \textbf{AIME} & \textbf{AIME25} & \textbf{M-Avg.} & \textbf{C-Avg.} & \textbf{Avg.} \\
\midrule
$|G|=8$   & 77.41 & 64.90 & 38.49 & 47.67 & 15.42 & 8.54  & 60.27 & 23.87 & 42.07 \\
$|G|=16$  & {77.49} & {67.21} & 38.97 & 49.17 & 18.13 & {10.63} & \textbf{61.22} & 25.97 & 43.60 \\
$|G|=32$  & 77.26 & 66.03 & {39.89} & {50.45} & {20.83} & 8.33  & 61.06 & \textbf{26.54} & \textbf{43.80} \\
\bottomrule
\end{tabular}
}
\end{table*}

\paragraph{Group Size Analysis.}
The effectiveness of \ours relies on group-based policy optimization, where the group size (i.e., the number of rollouts) can significantly influence contrastive learning outcomes.
A larger number of rollouts results in more non-trivial groups, providing denser reward signals.
Additionally, more rollouts offer a richer and more diverse set of positive and negative samples for contrastive learning.
We conduct experiments to evaluate the impact of varying the number of rollouts. The results are presented in Table~\ref{tab:more_group_size}.
These results indicate that increasing the number of rollouts generally leads to improved performance, highlighting the importance of group size in enhancing the effectiveness of contrastive rewards.
In the Comp-Avg. metric, as expected, the performance gradually improves with an increasing number of rollouts. Group size of 32 achieves the highest Comp-Avg. score of 26.54, indicating that larger groups provide more diverse and informative samples for the model to learn from.
We also conduct ablation studies on group size with different temperature settings, yielding similar results.
Additional results and details are provided in Appendix~\ref{sec:app_group_size}.


\begin{wraptable}{r}{0.5\textwidth}
\centering
\caption{Performance comparison on diverse models.}
\label{tab:more_models}

\resizebox{\linewidth}{!}{%
\begin{tabular}{llccc}
\toprule
\textbf{Model} & \textbf{Method} & \textbf{Math-Avg.} & \textbf{Comp-Avg.} & \textbf{Avg.} \\
\midrule
\multirow{3}{*}{DS-7B}
  & GRPO & 79.39 & 45.38 & 62.38 \\
  & +\ours & 79.28 & 46.54 & 62.91 \\
  & \cellcolor{gray!10}$\Delta$ & \cellcolor{neg}-0.11 & \cellcolor{pos}1.17 & \cellcolor{pos}0.53 \\
\midrule
\multirow{3}{*}{Llama-8B}
  & GRPO & 34.84 & 11.42 & 23.13 \\
  & +\ours & 37.10 & 11.79 & 24.44 \\
  & \cellcolor{gray!10}$\Delta$ & \cellcolor{pos}2.25 & \cellcolor{pos}0.37 & \cellcolor{pos}1.31 \\
\bottomrule
\end{tabular}
}
\end{wraptable}

\paragraph{Base Model Variations.}
To further validate the generalizability of our proposed method across different architectures, we conduct additional experiments using various base models, including DeepSeek-R1-Distill-Qwen-7B (DS-7B), and Llama3.1-8B (Llama-8B).
As presented in Table~\ref{tab:more_models}, we observe that both DS-7B and Llama-8B exhibit similar performance improvements upon integrating \ours.
Specifically, DS-7B shows an increase of +0.53 in average score, while Llama-8B achieves a notable gain of +1.31.
These results underscore the versatility and effectiveness of our approach in enhancing reasoning capabilities across diverse language model architectures.
Notably, there is a slight decrease of -0.11 in Math-Avg. for DS-7B.
We speculate that DS-7B has already undergone extensive fine-tuning on mathematical tasks, achieving a high level of performance.
When trained for the same number of epochs as other base models, it may experience overfitting, leading to a slight decrease in performance on certain tasks.
The detailed experimental results based on different models are presented in Appendix~\ref{sec:app_base_model}.

\section{Conclusions}

We introduced {\ours}, a contrastive learning augmented framework that generalizes RLVR beyond coarse, outcome-based supervision. Specifically, we employ a lightweight contrastive head with an InfoNCE objective to align the representations of successful reasoning trajectories while enforcing separation from erroneous ones. This design effectively distills the latent semantic consistency among successful rollouts into a dense contrastive reward, which complements standard verifier feedback. By reshaping the reward landscape from sparse and binary to informative and relational, \ours facilitates more stable policy optimization without requiring external process annotations or additional supervision. Extensive experiments across diverse mathematical benchmarks validate the efficacy of our approach: notably, consistent gains on symbolic, perturbed, and out-of-distribution tasks demonstrate that contrastive reward shaping fosters more robust and generalizable reasoning. Ultimately, this work highlights a promising paradigm for advancing RLVR, that leveraging the inherent relational structure among successful reasoning solutions as a principled and scalable learning signal.
 Beyond mathematical reasoning, \ours is broadly applicable to other structured domains like code generation and agent planning, paving the way for more reliable and generalized reasoning intelligence.   

\clearpage
\bibliography{iclr2026_conference}
\bibliographystyle{iclr2026_conference}

\newpage
\appendix
\section{Appendix}



\subsection{Implementation Details}
\label{sec:app_implementation_details}

\subsubsection{Baseline Implementations}
We reproduce GRPO, GSPO, DAPO, GMPO, using the public verl framework~\citep{sheng2024hybridflow}. 
All baselines share identical data, backbone initialization, sampling strategy, batch configuration, optimization strategy, and distributed training settings; the only differences arise from their respective policy objectives and advantage aggregation schemes. We summarize the key distinctions below.

\begin{table}[h]
\centering
\caption{Default Training Hyperparameters and Configuration on Track I and Track II.}
\label{tab:app_training_params}
\small
\begin{tabular}{llll}
\toprule
\textbf{Category} & \textbf{Parameter} & \textbf{Track I} & \textbf{Track II} \\
\midrule
\multicolumn{4}{l}{\textbf{General Settings}} \\
Total epochs & total\_epochs & 6 & 9 \\
Critic warmup epochs & critic\_warmup & 14 step(1 epoch) & 58 step(1 epoch) \\
\midrule
\multicolumn{4}{l}{\textbf{Data and Sequence Lengths}} \\
Training batch size & train\_batch\_size & 512 & 128 \\
PPO mini-batch size & ppo\_mini\_batch\_size & 256 & 128 \\
Micro batch size / GPU & micro\_batch\_size & 32 & 16 \\
Max prompt length & max\_prompt\_length & 512 & 512 \\
Max response length & max\_response\_length & 2048 & 4096 \\
Prompt overlength handling & filter\_overlong\_prompts & True & True \\
Truncation policy & truncation & error & error \\
\midrule
\multicolumn{4}{l}{\textbf{Actor and PPO Optimization}} \\
Actor learning rate & actor.optim.lr & $1\times10^{-6}$ & $1\times10^{-6}$ \\
Dynamic batch size & use\_dynamic\_bsz & True & True \\
Gradient checkpointing & enable\_gradient\_checkpointing & True & True \\
\midrule
\multicolumn{4}{l}{\textbf{Rollout and Inference}} \\
Rollout backend & rollout.name & sglang & sglang \\
Rollout samples ($n$) & rollout.n & 16 & 16 \\
Validation rollouts & val\_rollout & 1 & 16 \\
Sampling during eval & do\_sample & False & True \\
Temperature (eval) & temperature & 0.6 & 0.6 \\
Top-$p$ (eval) & top\_p & 0.95 & 0.95 \\
\midrule
\multicolumn{4}{l}{\textbf{Distributed Training}} \\
GPUs per node & n\_gpus\_per\_node & 8 & 8 \\
Number of nodes & nnodes & 1 & 1 \\
Tensor model parallel size & tensor\_model\_parallel\_size & 1 & 1 \\
FSDP parameter offload (actor) & fsdp.param\_offload & False & False \\
FSDP optimizer offload (actor) & fsdp.optimizer\_offload & False & False \\
FSDP parameter offload (ref) & ref.fsdp.param\_offload & True & False \\
\bottomrule
\end{tabular}
\end{table}

\paragraph{GRPO.}
GRPO follows a PPO-style clipped surrogate objective while incorporating an explicit KL penalty as a separate loss term. In our implementation, reward shaping does not include KL divergence, whereas an auxiliary KL loss with a fixed coefficient is enabled:
\[
\mathcal{L}_{\mathrm{GRPO}}
=
\mathbb{E}\!\left[ \mathrm{clip}(\rho_t, 1-\epsilon, 1+\epsilon) A_t \right]
-
\beta \, \mathrm{KL}(\pi \,\|\, \pi_{\mathrm{ref}}).
\]
The corresponding configuration sets\texttt{use\_kl\_loss=True}, and \texttt{kl\_loss\_coef=0.001}. Advantage aggregation is performed via token averaging, specified by \texttt{loss\_agg\_mode=token-mean}, with a symmetric clipping factor of \(\epsilon=0.2\).

\paragraph{GSPO.}
GSPO replaces token-wise advantages with group-wise preference signals and adopts a very narrow clipping interval, as recommended by the original paper~\citep{zheng2025gspo}. The loss employs a sequence-level aggregation.
In our setup, we configure \texttt{loss\_mode=gspo}, \texttt{loss\_agg\_mode=seq-mean-token-mean}, and use \(\mathrm{clip\_ratio\_low}=3\times10^{-4}\) and \(\mathrm{clip\_ratio\_high}=4\times10^{-4}\), aligning with the suggested hyperparameters.

\paragraph{DAPO.}
DAPO maintains a PPO-like update but introduces an asymmetric clipping interval:
\[
\mathcal{L}_{\mathrm{DAPO}}
=
\mathbb{E}\!\left[
\mathrm{clip}(\rho_t, 1-\epsilon_{l}, 1+\epsilon_{h}) A_t
\right]
\]
We set \(\epsilon_{l}=0.2\) and \(\epsilon_{h}=0.28\), following the original recommendations~\citep{yu2025dapo}.
Both KL reward shaping and KL loss regularization are disabled (\texttt{use\_kl\_in\_reward=False} and \texttt{use\_kl\_loss=False}).

\paragraph{GMPO.}
GMPO adopts a geometric-mean policy objective in place of additive advantage accumulation,
which promotes multiplicative update dynamics and mitigates domination by outlier advantages. 
GMPO uses a single symmetric clipping factor with \(\epsilon=0.4\), enforced by setting \texttt{clip\_ratio\_low=clip\_ratio\_high=0.4}, and applies token-level advantage aggregation via \texttt{loss\_agg\_mode=token-mean}.

\paragraph{Shared Training Protocol.}
All baselines are trained under a unified protocol, sharing identical hyperparameters for learning rate, rollout sampling configuration (temperature, top-p, number of rollouts), batch size, sequence length, and distributed memory optimization settings. 
This ensures that performance differences can be attributed solely to the distinct policy objectives and advantage aggregation methods employed by each algorithm.
The complete set of shared training hyperparameters is detailed in Table~\ref{tab:app_training_params}.

\paragraph{Contrastive Head Parameters}
We detail the architecture and hyperparameters of the Contrastive Head in Table~\ref{tab:app_training_params_head}.

\begin{table}[h]
\centering
\caption{Default Contrastive Head Parameters on Track I and Track II.}
\label{tab:app_training_params_head}
\small
\begin{tabular}{lll}
\toprule
\textbf{Parameter} & \textbf{Track I} & \textbf{Track II} \\
\midrule
\multicolumn{3}{l}{\textbf{Contrastive LM Head Settings}} \\
Loss type & infonce\_loss & infonce\_loss \\
Distance measure & cosine\_distance & cosine\_distance \\
Head type & linear & linear \\
Output size & 512 & 2048 \\
Contrastive temperature & 0.05 & 0.05 \\
Learning rate & 0.001 & 0.001 \\
Weight decay & 0.01 & 0.01 \\
$\lambda$ & 0.2 & 0.2 \\
Negative Selection & group\_level & group\_level \\
\bottomrule
\end{tabular}
\end{table}

\subsubsection{Prompt Construction}
\label{sec:app_prompt}

For all datasets, each training and evaluation example is converted into a unified
instruction-following format.
Given an original problem statement $q$ from dataset $d$, we construct the user prompt as
\begin{equation}
\texttt{Prompt}(q, d) =
q \; \Vert \; \texttt{Instruction}(d),
\end{equation}
where $\Vert$ denotes string concatenation with a newline separator.

The dataset-specific instruction $\texttt{Instruction}(d)$ is appended to explicitly
encourage step-by-step reasoning and constrain the answer format.
All prompts are assigned the \texttt{user} role and no system or assistant messages are used.

\paragraph{Instruction Templates.}
For most mathematical reasoning datasets, including GSM8K, MATH, AMC, AIME, and their variants,
we use the following instruction:
\begin{quote}
\texttt{Let's think step by step and output the final answer within \textbackslash boxed\{\}.}
\end{quote}

For multiple-choice datasets, the instruction additionally specifies the expected answer type.
For example, for MMLU:
\begin{quote}
\texttt{Let's think step by step and output the final answer (eg, A, B, C, D) within \textbackslash boxed\{\}.}
\end{quote}

We summarize the complete set of dataset-specific prompt instructions in Table~\ref{tab:app_prompt_template}.

\begin{table}[ht]
\centering
\caption{Prompt Construction and Instruction Templates}
\label{tab:app_prompt_template}
\small
\begin{tabularx}{\linewidth}{lX}
\toprule
\textbf{Dataset} & \textbf{Instruction Template} \\
\midrule
GSM8K / GSM-P1 / GSM-P2 / GSM-Symbolic &
Let's think step by step and output the final answer within \textbackslash boxed\{\}. \\
\midrule
MATH / AMC / AIME / AIME2025 &
Let's think step by step and output the final answer within \textbackslash boxed\{\}. \\
\midrule
TheoremQA &
Let's think step by step and output the final answer within \textbackslash boxed\{\}. \\
\midrule
MMLU &
Let's think step by step and output the final answer (eg, A, B, C, D) within \textbackslash boxed\{\}. \\
\midrule
TruthfulQA &
Let's think step by step and output the final answer (eg, A, B, C, D, ...) within \textbackslash boxed\{\}. \\
\midrule
CommonsenseQA &
Let's think step by step and output the final answer (eg, A, B, C, D, E) within \textbackslash boxed\{\}. \\
\bottomrule
\end{tabularx}
\end{table}

\subsubsection{Verifiable Reward Models}
\label{sec:app_reward_model}
Following prior work~\citep{shao2024deepseekmath,aggarwal2025l1}, we employ rule-based verifiable reward models to compute rewards based on the final answers generated by the LLM.
Specifically, we extract the final answer enclosed within the \texttt{\textbackslash boxed\{\}} delimiters. 
This extracted answer, denoted as \( \text{LLM}_{\text{pred}} \), is then compared against the ground truth answer \( \text{GT} \) to determine correctness.
The reward \( r \) is 0 for incorrect answers and 1 for correct answers.

\subsection{Contrastive Head Representation and Ablation}
\label{sec:app_contrastive_head}
To gain insights into the representations learned by the contrastive head, we conduct a qualitative analysis of the semantic embeddings it produces. We randomly sample a set of responses generated during training and compute their embeddings using the contrastive head. 

\begin{figure}[!ht]
    \centering
    \includegraphics[width=\linewidth]{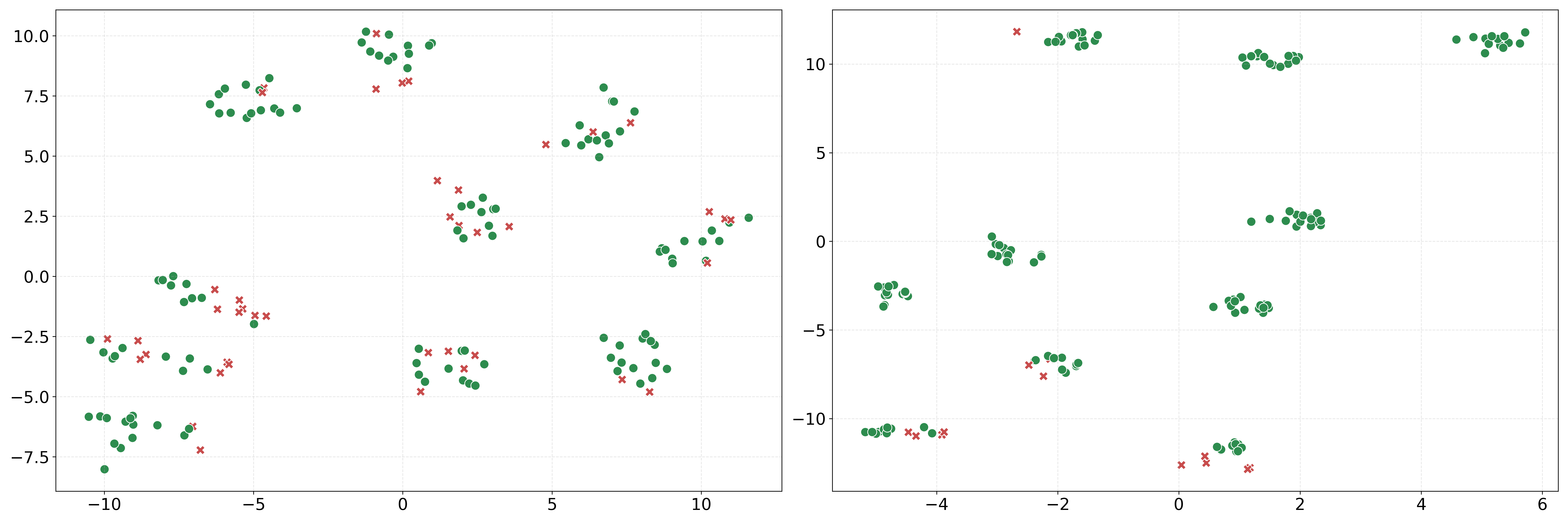}
    \caption{The t-SNE visualization of semantic embeddings produced by the contrastive head at the start of training (left) and after $3$ epochs of training (right).
    Green points represent embeddings from correct rollouts, while red points correspond to incorrect ones. 
    After training, correct responses cluster closely together, forming more distinct group clusters. Within these clusters, correct and incorrect responses also exhibit some separation.
    }
    \label{fig:embedding_tsne}
\end{figure}

As illustrated in Figure~\ref{fig:embedding_tsne}, the embeddings exhibit no clear separation: correct and incorrect trajectories are mixed and lack coherent structure at the start of training. 
After contrastive training, the embeddings exhibit a clear clustering structure. 
Embeddings corresponding to correct rollouts tend to cluster closely together, while those from incorrect rollouts are relatively separated.
This structure aligns with the goal of the contrastive reward, which encourages higher-quality solutions to be close in representation space while pushing away suboptimal ones. 
Importantly, this signal is more fine-grained than binary reward feedback, since embeddings reflect relative similarity of trajectories even when they share the same correctness outcome.

This observation further suggests that the contrastive head is not merely serving as an auxiliary regression module, but is shaping a semantic manifold for mathematical reasoning trajectories. The emergence of such a manifold provides a meaningful inductive bias for group-based policy optimization: candidate solutions that share intermediate reasoning states or substructures are more likely to benefit from shared credit assignment.

To validate that the learned geometry is not solely inherited from the pretrained LLM backbone, we conduct an ablation where the contrastive head is fixed and not updated during training (denoted as \ours-fixed). 

\begin{table*}[htbp]
\centering
\caption{Performance comparison of fixed-head ablation on Track I.
\ours-fixed shows performance drops compared to \ours across most benchmarks.
}
\label{tab:app_fixedhead_complete_1}
\resizebox{\textwidth}{!}{%
\begin{tabular}{lccccccccccc}
\toprule
\textbf{Methods} & \textbf{GSM8K} & \textbf{SYM} & \textbf{P1} & \textbf{P2} & \textbf{cQA} & \textbf{tQA} & \textbf{thrmQA} & \textbf{MMLU} & \textbf{M-Avg.} & \textbf{G-Avg.} & \textbf{Avg.} \\ \midrule
\ours & 88.02 & 84.62 & 74.60 & 54.16 & 76.90 & 31.52 & 27.58 & 68.64 & 75.35 & 51.16 & 63.26 \\
\ours-fixed & 88.63 & 85.12 & 73.48 & 51.68 & 75.59 & 30.11 & 26.77 & 68.49 & 74.73 & 50.24 & 62.48 \\
$\Delta$ &
\cellcolor{pos}0.61 &
\cellcolor{pos}0.50 &
\cellcolor{neg}-1.12 &
\cellcolor{neg}-2.48 &
\cellcolor{neg}-1.31 &
\cellcolor{neg}-1.41 &
\cellcolor{neg}-0.80 &
\cellcolor{neg}-0.15 &
\cellcolor{neg}-0.62 &
\cellcolor{neg}-0.92 &
\cellcolor{neg}-0.77 \\
\bottomrule
\end{tabular}%
}
\end{table*}

\begin{table*}[htbp]
\centering
\caption{Performance comparison of fixed-head ablation on Track II.
\ours-fixed shows performance drops compared to \ours across most benchmarks.
}
\label{tab:app_fixedhead_complete_2}
\resizebox{\textwidth}{!}{%
\begin{tabular}{lccccccccc}
\toprule
\textbf{Methods} & \textbf{Math} & \textbf{Math\_S} & \textbf{Math\_H} & \textbf{AMC} & \textbf{AIME} & \textbf{AIME25} & \textbf{M-Avg.} & \textbf{C-Avg.} & \textbf{Avg.} \\
\midrule
\ours & 77.49 & 67.21 & 38.97 & 49.17 & 18.13 & 10.63 & 61.22 & 25.97 & 43.60 \\
\ours-fixed & 76.76 & 64.83 & 38.99 & 49.77 & 17.92 & 7.50 & 60.19 & 25.06 & 42.63 \\
$\Delta$ &
\cellcolor{neg}-0.73 &
\cellcolor{neg}-2.38 &
\cellcolor{pos}0.02 &
\cellcolor{pos}0.60 &
\cellcolor{neg}-0.21 &
\cellcolor{neg}-3.13 &
\cellcolor{neg}-1.03 &
\cellcolor{neg}-0.91 &
\cellcolor{neg}-0.97 \\
\bottomrule
\end{tabular}
}
\end{table*}

As shown in Table~\ref{tab:app_fixedhead_complete_1}, freezing the head leads to consistent performance drops across all three average metrics, particularly on distribution-shifted and perturbed datasets (P1, P2), with decreases of -1.12 and -2.48, respectively.
Table~\ref{tab:app_fixedhead_complete_2} shows a similar trend, where fixing the head results in overall performance degradation, with decreases of -1.03, -0.91, and -0.97 on M-Avg, C-Avg, and Avg, respectively.
These results highlight that the mapping learned by the contrastive head is non-trivial and contributes useful inductive structure.

Together, these analyses indicate that the contrastive head learns a structured embedding space that encodes trajectory quality, enabling more informative credit assignment and yielding improved generalization, particularly under distribution shifts and perturbations.

\begin{table*}[ht]
\centering
\caption{Performance comparison of Contrastive Loss Variants on Track I. 
The improvements ($\Delta$) over base method(GRPO) are indicated with \cellcolor{pos}positive (green) and \cellcolor{neg}negative (red) colors.
}
\label{tab:app_loss_track1}
\resizebox{\textwidth}{!}{%
\begin{tabular}{l *{11}{S[table-format=2.2, table-column-width=0.98cm]}}
\toprule
\textbf{Methods} & \textbf{GSM8K} & \textbf{SYM} & \textbf{P1} & \textbf{P2} & \textbf{cQA} & \textbf{tQA} & \textbf{thrmQA} & \textbf{MMLU} & \textbf{M-Avg.} & \textbf{G-Avg.} & \textbf{Avg.} \\
\midrule
GRPO & 87.79 & 84.04 & 73.12 & 50.80 & 74.94 & 30.48 & 27.71 & 68.19 & 73.94 & 50.33 & 62.13 \\
\ours-InfoNCE & 88.02 & 84.62 & 74.60 & 54.16 & 76.90 & 31.52 & 27.58 & 68.64 & 75.35 & 51.16 & 63.26 \\
\cellcolor{gray!10}$\Delta$ 
& \cellcolor{pos}0.23 & \cellcolor{pos}0.58 & \cellcolor{pos}1.48 & \cellcolor{pos}3.36 & \cellcolor{pos}1.97 & \cellcolor{pos}1.04 & \cellcolor{neg}-0.13 & \cellcolor{pos}0.46 & \cellcolor{pos}1.41 & \cellcolor{pos}0.83 & \cellcolor{pos}1.12 \\
\ours-SupCon & 87.79 & 85.04 & 74.00 & 51.12 & 77.07 & 30.23 & 27.04 & 69.25 & 74.49 & 50.90 & 62.69 \\
\cellcolor{gray!10}$\Delta$
& \cellcolor{gray!10}0.00 & \cellcolor{pos}1.00 & \cellcolor{pos}0.88 & \cellcolor{pos}0.32 & \cellcolor{pos}2.13 & \cellcolor{neg}-0.24 & \cellcolor{neg}-0.67 & \cellcolor{pos}1.06 & \cellcolor{pos}0.55 & \cellcolor{pos}0.57 & \cellcolor{pos}0.56 \\
\ours-SoftNN & 87.11 & 84.70 & 74.06 & 53.92 & 75.27 & 29.99 & 27.31 & 68.59 & 74.95 & 50.29 & 62.62 \\
\cellcolor{gray!10}$\Delta$
& \cellcolor{neg}-0.68 & \cellcolor{pos}0.66 & \cellcolor{pos}0.94 & \cellcolor{pos}3.12 & \cellcolor{pos}0.33 & \cellcolor{neg}-0.49 & \cellcolor{neg}-0.40 & \cellcolor{pos}0.41 & \cellcolor{pos}1.01 & \cellcolor{neg}-0.04 & \cellcolor{pos}0.48 \\
\bottomrule
\end{tabular}
}
\end{table*}

\begin{table*}[ht]
\centering
\caption{Performance comparison of Contrastive Loss Variants on Track II. 
The improvements ($\Delta$) over base method(GRPO) are indicated with \cellcolor{pos}positive (green) and \cellcolor{neg}negative (red) colors.
}
\label{tab:app_loss_track2}
\resizebox{\textwidth}{!}{%
\begin{tabular}{l *{9}{S[table-format=2.2, table-column-width=1.2cm]}}
\toprule
\textbf{Methods} & \textbf{Math} & \textbf{Math\_S} & \textbf{Math\_H} & \textbf{AMC} & \textbf{AIME} & \textbf{AIME25} & \textbf{M-Avg.} & \textbf{C-Avg.} & \textbf{Avg.} \\
\midrule
GRPO & 76.46 & 64.83 & 37.86 & 46.84 & 17.92 & 9.58 & 59.72 & 24.78 & 42.25 \\
\ours-InfoNCE & 77.49 & 67.21 & 38.97 & 49.17 & 18.13 & 10.63 & 61.22 & 25.97 & 43.60 \\
\cellcolor{gray!10}$\Delta$
& \cellcolor{pos}1.03 & \cellcolor{pos}2.38 & \cellcolor{pos}1.11 & \cellcolor{pos}2.33 & \cellcolor{pos}0.21 & \cellcolor{pos}1.04 & \cellcolor{pos}1.50 & \cellcolor{pos}1.19 & \cellcolor{pos}1.35 \\
\ours-SupCon & 78.18 & 65.90 & 38.97 & 48.42 & 17.71 & 6.88 & 61.01 & 24.33 & 42.67 \\
\cellcolor{gray!10}$\Delta$
& \cellcolor{pos}1.72 & \cellcolor{pos}1.06 & \cellcolor{pos}1.11 & \cellcolor{pos}1.58 & \cellcolor{neg}-0.21 & \cellcolor{neg}-2.71 & \cellcolor{pos}1.30 & \cellcolor{neg}-0.45 & \cellcolor{pos}0.43 \\
\ours-SoftNN & 78.17 & 65.78 & 38.02 & 49.85 & 20.21 & 9.58 & 60.66 & 26.55 & 43.60 \\
\cellcolor{gray!10}$\Delta$
& \cellcolor{pos}1.71 & \cellcolor{pos}0.95 & \cellcolor{pos}0.16 & \cellcolor{pos}3.01 & \cellcolor{pos}2.29 & 0.00 & \cellcolor{pos}0.94 & \cellcolor{pos}1.77 & \cellcolor{pos}1.35 \\
\bottomrule
\end{tabular}
}
\end{table*}

\subsection{Contrastive Loss Variants}
\label{sec:app_more_losses}
We have also implemented additional loss functions for the Contrastive LM Head, including Soft Nearest Neighbor (SoftNN) loss and Supervised Contrastive (SupCon) loss. We refer to the corresponding methods as \ours-SoftNN and \ours-SupCon, respectively.
The main experimental results for these methods on GSM8K and MATH datasets are presented in Tables~\ref{tab:app_loss_track1} and~\ref{tab:app_loss_track2}.

One observation is that SoftNN exhibits a performance drop on G-Avg.
To further investigate this phenomenon, we conduct additional ablation experiments using SoftNN loss across different base methods.
The complete experimental results are presented in Table~\ref{tab:app_softnn_gsm}.

\begin{table*}[t]
\centering
\caption{Performance comparison of different RLVR methods.
\ours-SoftNN improves the average performance based on four baseline methods, with the main improvements in M-Avg. tasks, while the improvements in G-Avg. tasks are limited, and even some decreases.
}
\label{tab:app_softnn_gsm}
\resizebox{\textwidth}{!}{%
\begin{tabular}{l *{11}{S[table-format=2.2, table-column-width=0.98cm]}}
\toprule
\textbf{Methods} & \textbf{GSM} & \textbf{SYM} & \textbf{P1} & \textbf{P2} & \textbf{cQA} & \textbf{tQA} & \textbf{thrmQA} & \textbf{MMLU} & \textbf{M-Avg.} & \textbf{G-Avg.} & \textbf{Avg.} \\ \midrule
Base & 86.05 & 82.40 & 70.90 & 47.96 & 73.63 & 29.99 & 24.90 & 68.09 & 71.83 & 49.15 & 60.49 \\
GRPO & 87.79 & 84.04 & 73.12 & 50.80 & 74.94 & 30.48 & 27.71 & 68.19 & 73.94 & 50.33 & 62.13 \\
\ours-SoftNN & 87.11 & 84.70 & 74.06 & 53.92 & 75.27 & 29.99 & 27.31 & 68.59 & 74.95 & 50.29 & 62.62 \\
\cellcolor{gray!10}$\Delta$ & \cellcolor{neg}-0.68 & \cellcolor{pos}0.66 & \cellcolor{pos}0.94 & \cellcolor{pos}3.12 & \cellcolor{pos}0.33 & \cellcolor{neg}-0.49 & \cellcolor{neg}-0.40 & \cellcolor{pos}0.41 & \cellcolor{pos}1.01 & \cellcolor{neg}-0.04 & \cellcolor{pos}0.48 \\
\hline
GSPO & 87.19 & 83.94 & 73.28 & 51.36 & 74.86 & 30.23 & 27.38 & 67.68 & 73.94 & 50.04 & 61.99 \\
\ours-SoftNN & 87.87 & 83.92 & 75.14 & 51.60 & 76.33 & 30.97 & 27.64 & 68.44 & 74.63 & 50.85 & 62.74 \\
\cellcolor{gray!10}$\Delta$ & \cellcolor{pos}0.68 & \cellcolor{neg}-0.02 & \cellcolor{pos}1.86 & \cellcolor{pos}0.24 & \cellcolor{pos}1.47 & \cellcolor{pos}0.73 & \cellcolor{pos}0.27 & \cellcolor{pos}0.76 & \cellcolor{pos}0.69 & \cellcolor{pos}0.81 & \cellcolor{pos}0.75 \\
\hline
DAPO & 87.41 & 84.56 & 74.70 & 52.32 & 76.00 & 31.58 & 27.11 & 69.00 & 74.75 & 50.92 & 62.84 \\
\ours-SoftNN & 88.25 & 84.48 & 76.10 & 53.52 & 75.84 & 31.82 & 27.24 & 67.68 & 75.59 & 50.65 & 63.12 \\
\cellcolor{gray!10}$\Delta$ & \cellcolor{pos}0.83 & \cellcolor{neg}-0.08 & \cellcolor{pos}1.40 & \cellcolor{pos}1.20 & \cellcolor{neg}-0.16 & \cellcolor{pos}0.24 & \cellcolor{pos}0.13 & \cellcolor{neg}-1.32 & \cellcolor{pos}0.84 & \cellcolor{neg}-0.28 & \cellcolor{pos}0.28 \\
\hline
GMPO & 86.81 & 84.02 & 73.90 & 49.80 & 76.33 & 30.97 & 26.57 & 68.49 & 73.63 & 50.59 & 62.11 \\
\ours-SoftNN & 87.87 & 84.14 & 73.94 & 51.32 & 75.84 & 31.52 & 26.24 & 67.63 & 74.32 & 50.31 & 62.31 \\
\cellcolor{gray!10}$\Delta$ & \cellcolor{pos}1.06 & \cellcolor{pos}0.12 & \cellcolor{pos}0.04 & \cellcolor{pos}1.52 & \cellcolor{neg}-0.49 & \cellcolor{pos}0.55 & \cellcolor{neg}-0.33 & \cellcolor{neg}-0.86 & \cellcolor{pos}0.69 & \cellcolor{neg}-0.28 & \cellcolor{pos}0.20 \\
\bottomrule
\end{tabular}%
}
\end{table*}

\subsection{Temperature Results}
\label{sec:app_temperature}
\begin{figure}[t]
    \centering
    \includegraphics[width=0.6\linewidth]{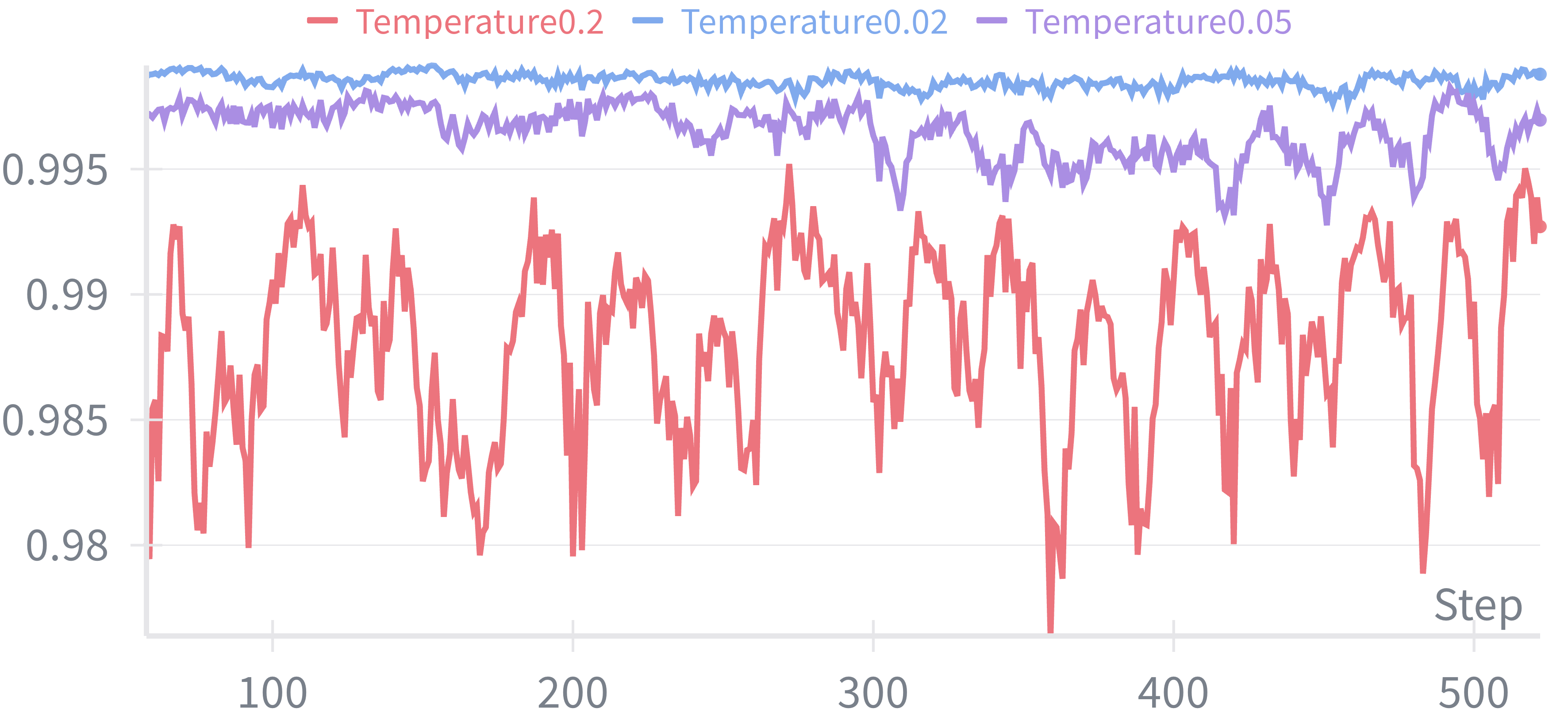}
    \caption{The plot illustrates the average cosine similarity of positive pairs within the same group during training, under different temperature settings. We observe that higher temperature values lead to greater fluctuations in the similarity of positive pairs.}
    \label{fig:embeddings_temperature}
\end{figure}

The complete experimental results across different temperature settings are presented in Table~\ref{tab:app_temp_gsm} and~\ref{tab:app_temp_math}.
Figure~\ref{fig:embeddings_temperature} visualizes the average cosine similarity of positive pairs within the same group during training, under different temperature settings.
The plot illustrates that higher temperature values lead to greater fluctuations in the similarity of positive pairs, indicating increased instability in distinguishing positive and negative examples.

\begin{table*}[t]
\centering
\caption{Performance comparison of different temperature settings on Track I.}
\label{tab:app_temp_gsm}
\resizebox{\textwidth}{!}{%
\begin{tabular}{l *{11}{S[table-format=2.2, table-column-width=0.98cm]}}
\toprule
\textbf{$\tau$} & \textbf{GSM8K} & \textbf{SYM} & \textbf{P1} & \textbf{P2} & \textbf{cQA} & \textbf{tQA} & \textbf{thrmQA} & \textbf{MMLU} & \textbf{M-Avg.} & \textbf{G-Avg.} & \textbf{Avg.} \\
\midrule
0.2  & 87.34 & 84.58 & 73.42 & 50.48 & \textbf{76.33} & \textbf{32.74} & \textbf{29.38} & 67.73 & 73.95 & \textbf{51.55} & 62.75 \\
0.1  & \textbf{88.02} & 85.38 & 74.50 & 49.64 & 75.92 & 31.95 & 27.91 & 68.14 & 74.39 & 50.98 & 62.68 \\
0.05 & 87.41 & \textbf{85.44} & 74.86 & 53.16 & 75.18 & 30.42 & 27.71 & \textbf{68.79} & 75.22 & 50.53 & \textbf{62.87} \\
0.02 & 87.26 & 85.00 & \textbf{75.64} & \textbf{54.00} & 75.10 & 30.17 & 27.31 & 68.34 & \textbf{75.48} & 50.23 & 62.85 \\
\bottomrule
\end{tabular}
}
\end{table*}

\begin{table*}[htbp]
\centering
\caption{Performance comparison of different temperature settings on Track II.}
\label{tab:app_temp_math}
\resizebox{\textwidth}{!}{%
\begin{tabular}{l *{9}{S[table-format=2.2, table-column-width=1.2cm]}}
\toprule
\textbf{$\tau$} & \textbf{Math} & \textbf{Math\_S} & \textbf{Math\_H} & \textbf{AMC} & \textbf{AIME} & \textbf{AIME25} & \textbf{M-Avg.} & \textbf{C-Avg.} & \textbf{Avg.} \\
\midrule
0.2   & 77.13 & 66.12 & 36.44 & 47.67 & 17.92 & 7.50 & 59.90 & 24.36 & 42.13 \\
0.1   & 76.90 & 65.01 & 37.36 & \textbf{51.05} & \textbf{18.75} & \textbf{10.63} & 59.76 & \textbf{26.81} & 43.28 \\
0.05  & 77.18 & 67.14 & \textbf{39.53} & 48.72 & 16.04 & 9.79 & \textbf{61.28} & 24.85 & 43.07 \\
0.02  & \textbf{77.49} & \textbf{67.21} & 38.97 & 49.17 & 18.13 & \textbf{10.63} & 61.22 & 25.97 & \textbf{43.60} \\
0.015 & 76.79 & 66.42 & 38.72 & 50.15 & 17.71 & 7.71 & 60.64 & 25.19 & 42.91 \\
\bottomrule
\end{tabular}
}
\end{table*}

\subsection{Group Size Results}
\label{sec:app_group_size}
The complete experimental results across different group size settings are presented in Table~\ref{tab:app_group_size}.
From the results, we observe that larger group sizes consistently lead to better performance improvements across different temperature settings.
This trend suggests that increasing the number of candidate solutions within each group enhances the contrastive signal, providing richer information for credit assignment and ultimately improving the model's reasoning capabilities.

\begin{table*}[t]
\centering
\caption{Performance comparison of different group size.}
\label{tab:app_group_size}
\resizebox{\textwidth}{!}{%
\begin{tabular}{l l *{9}{S[table-format=2.2, table-column-width=1.2cm]}}
\toprule
\textbf{$\tau$} & \textbf{GroupSize} & \textbf{Math} & \textbf{Math\_S} & \textbf{Math\_H} & \textbf{AMC} & \textbf{AIME} & \textbf{AIME25} & \textbf{M-Avg.} & \textbf{C-Avg.} & \textbf{Avg.} \\
\midrule
\multirow{3}{*}{0.05} & $|G|=8$  & 76.56 & 64.92 & 38.09 & 46.69 & 17.50 & \textbf{10.00} & 59.86 & 24.73 & 42.29 \\
                      & $|G|=16$ & 77.18 & \textbf{67.14} & \textbf{39.53} & 48.72 & 16.04 & 9.79 & \textbf{61.28} & 24.85 & 43.07 \\
                      & $|G|=32$ & \textbf{77.92} & 64.90 & 39.37 & \textbf{49.55} & \textbf{21.04} & 7.08 & 60.73 & \textbf{25.89} & \textbf{43.31} \\
\midrule
\multirow{3}{*}{0.02} & $|G|=8$   & 77.41 & 64.90 & 38.49 & 47.67 & 15.42 & 8.54  & 60.27 & 23.87 & 42.07 \\
                      & $|G|=16$  & \textbf{77.49} & \textbf{67.21} & 38.97 & 49.17 & 18.13 & \textbf{10.63} & \textbf{61.22} & 25.97 & 43.60 \\
                      & $|G|=32$  & 77.26 & 66.03 & \textbf{39.89} & \textbf{50.45} & \textbf{20.83} & 8.33  & 61.06 & \textbf{26.54} & \textbf{43.80} \\
\bottomrule
\end{tabular}
}
\end{table*}


\begin{table*}[t]
\centering
\caption{Performance comparison on various base models.
}
\label{tab:app_more_models}
\resizebox{\textwidth}{!}{%
\begin{tabular}{lccccccccc}
\toprule
\textbf{Methods} & \textbf{Math} & \textbf{Math\_S} & \textbf{Math\_H} & \textbf{AMC} & \textbf{AIME} & \textbf{AIME25} & \textbf{M-Avg.} & \textbf{C-Avg.} & \textbf{Avg.} \\
\midrule
\multicolumn{10}{l}{\textbf{DeepSeek-R1-Distill-Qwen-7B}} \\
GRPO & 89.70 & 83.98 & 64.48 & 72.59 & 36.88 & 26.67 & 79.39 & 45.38 & 62.38 \\
\ours & 90.08 & 83.94 & 63.83 & 73.80 & 39.38 & 26.46 & 79.28 & 46.54 & 62.91 \\
$\Delta$
& 0.38 & -0.05 & -0.65 & 1.20 & 2.50 & -0.21 & -0.11 & 1.17 & 0.53 \\
\midrule
\multicolumn{10}{l}{\textbf{Llama3.1-8B-Instruct}} \\
GRPO & 52.04 & 40.10 & 12.39 & 23.64 & 10.63 & 0.00 & 34.84 & 11.42 & 23.13 \\
\ours & 54.14 & 42.87 & 14.28 & 25.38 & 9.79 & 0.21 & 37.10 & 11.79 & 24.44 \\
$\Delta$
& 2.10 & 2.76 & 1.90 & 1.73 & -0.83 & 0.21 & 2.25 & 0.37 & 1.31 \\
\bottomrule
\end{tabular}
}
\end{table*}

\subsection{Base Model Variants}
\label{sec:app_base_model}
Beyond Qwen2.5, we also validate the effectiveness of \ours on other base models, including DS-7B and Llama-8B. The results are presented in Table~\ref{tab:app_more_models}.

\end{document}